\documentclass[twocolumn]{biophys-new}
\usepackage[utf8]{inputenc}
\usepackage{graphicx}
\usepackage[colorlinks,allcolors=cyan!70!black]{hyperref}
\usepackage{xcolor}

\usepackage{amsthm}
\usepackage{multirow}
\usepackage{multicol}
\usepackage{array}
\newtheorem{theorem}{Theorem}
\newtheorem{proposition}[theorem]{Proposition}
\theoremstyle{definition}\newtheorem{definition}{Definition}

\newtagform{noparen}{}{}
\usetagform{noparen}

\newcommand{\RR}{\mathbb{R}} %real numbers
\DeclareMathOperator*{\mae}{\it median}
\DeclareMathOperator*{\argmin}{arg\,min}

\title{A Topological Deep Learning Framework for Neural Spike Decoding}
\runningtitle{A Topological Deep Learning Framework for Neural Spike Decoding} %% For page header

\author[1]{Edward C.~Mitchell}
\author[1]{Brittany Story}
\author[2]{David Boothe}
\author[2,3]{Piotr J.~Franaszczuk}
\author[1,*]{Vasileios Maroulas}
\runningauthor{Micthell, Story, Boothe, Franaszczuk, Maroulas} %% For page header

\affil[1]{University of Tennessee Knoxville, Knoxville, Tennessee}
\affil[2]{Army Research Lab, Aberdeen, Maryland}
\affil[3]{Johns Hopkins University, Baltimore, Maryland}

\corrauthor[*]{vmaroula@utk.edu}

\papertype{Article}

\begin{document}
\begin{frontmatter}
%%%%%%%%%%%%
\begin{abstract}
The brain's spatial orientation system uses different neuron ensembles to aid in environment-based navigation.
Two of the ways brains encode spatial information is through head direction cells and grid cells.
Brains use head direction cells to determine orientation whereas grid cells consist of layers of decked neurons that overlay to provide environment-based navigation.
These neurons fire in ensembles where several neurons fire at once to activate a single head direction or grid.
We want to capture this firing structure and use it to decode head direction grid cell data.
Understanding, representing, and decoding these neural structures requires models that encompass higher order connectivity, more than the 1-dimensional connectivity that traditional graph-based models provide. 
To that end, in this work, we develop a topological deep learning framework for neural spike train decoding. 
Our framework combines unsupervised simplicial complex discovery with the power of deep learning via a new architecture we develop herein called a simplicial convolutional recurrent neural network.
Simplicial complexes, topological spaces that use not only vertices and edges but also higher-dimensional objects, naturally generalize graphs and capture more than just pairwise relationships.
Additionally, this approach does not require prior knowledge of the neural activity beyond spike counts, which removes the need for similarity measurements.
The effectiveness and versatility of the simplicial convolutional neural network is demonstrated on head direction and trajectory prediction via head direction and grid cell datasets. 
\end{abstract}

%%%%%%%%%%%%
\begin{sigstatement}
% Each manuscript must also have a statement of significance or no more than 120 words. 
We propose the simplicial convolutional recurrent neural network (SCRNN) as a method for decoding navigation cell spike trains. 
The SCRNN utilizes simplicial complexes, a tool from computational topology that captures higher order connectivity, paired with a recurrent neural network to decode head direction and grid cell spiking data.
The simplicial convolutional layer captures the firing structure of the neurons and utilizes the underlying connectivity as an input into the backend recurrent neural network.
We compared the median absolute error of the optimized SCRNN to those of three optimized traditional neural networks, a feedforward neural network, a recurrent neural network, and a graph neural network.
We conclude the SCRNN is able to predict head direction and grid activation better than three other traditional neural networks.
\end{sigstatement}
\end{frontmatter}

%%%%%%%%%%%%%%%%%%%%%%%%%%%%%%%%%%%%%%%%%%%%%%%%

% \textbf{Keywords:} simplicial complex, topological data analysis, grid cells, head direction cells, automated navigation 
%%%%%%%%%%%%%%%%%%%%%%%%%%%%%%%%%%%%%%%%%%%%%%%%
%%%%%%%%%%%%%%%%%%%%%%%%%%%%%%%%%%%%%%%%%%%%%%%%
\section{Introduction}
\label{sec:intro}

Neurophysiological recording techniques have produced simultaneous recordings from increased numbers of neurons, both in vitro and in vivo, allowing for access to the activity of the hundreds of neurons required to encode certain variables~\cite{Gardner2022, Yoshida2020naturalimages, Jun2017, Steinmetz2021}. 
This makes efficient algorithms for decoding the information content from neural spike trains of increasing interest.
Neural decoding can help provide insight into the function and significance of individual neurons or even entire regions of the brain~\cite{Glaser2019}.
Additionally, neural decoding provides a foundation for new machine learning algorithms which leverage the mammalian brain structure. 
Utilizing lower dimensional structure is one way the mammalian brain brings efficiency into neural data processing. 
Head direction cells and grid cells are two types of brain cells recorded in a quantity that allows for the analysis of their functional connectivity and structure of their population activity ~\cite{Peyrache2015, Gardner2022}.
The activity of head direction cells has been shown to live on a circle \cite{Chaudhuri2019}, whereas the activity of a module of grid cells lives on a torus~\cite{Gardner2022}. 
Hence, algorithmic tools that capture and utilize the inherent structure in the data are well equipped to decode neural spiking data. 

\begin{figure*}[t!]
    \centering
    \includegraphics[width=0.95\textwidth]{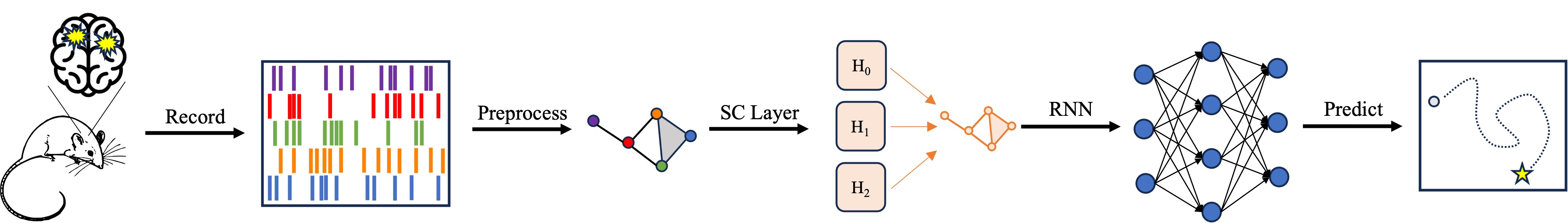}
    \caption{The framework for the SCRNN. 
    First, captured neural spiking data is recorded in a spike matrix. 
    Next, the data is converted into a simplicial complex for a series of time windows. 
    Then, each simplicial complex is fed into the simplicial convolutional layers.
    After flattening the simplicial information into a single vector, the vector is fed through a RNN, which predicts the location (or head direction) of the mouse based on the neural firing data.}
    \label{fig:intro}
\end{figure*}

Decoding methods typically employ statistical or deep learning based models since one may view them as a regression problem where we learn the relationship between the independent spike trains and the decoded dependent variable. 
Statistical methods like, but not limited to, linear regression, Bayesian reconstruction, and Kalman filtering are utilized for their interpretability and relatively low computational cost~\cite{Glaser2020, Peyrache2015, Xu2019}.
On the other hand, deep learning for neural decoding is a rapidly growing field due to neural networks' observed success at time-series tasks like sequence prediction and their ability to generalize beyond training data~\cite{Xu2019, Krizhevsky2012ImageClass, LeCun2015DeepLearning, Szabo2022}. 
Neural networks have outperformed statistical methods at decoding head direction and two-dimensional, environment-based position from neural recordings of head direction (HD) cells and place cells, respectively~\cite{Xu2019, Frey2019, Tampuu2019}. 
Deep learning's superior decoding performance has been observed for a variety of network architectures including recurrent (RNNs)~\cite{Elman1990, Rumelhart1986}, fully-connected feed forward neural networks (FFNNs), and convolutional neural networks (CNNs)~\cite{LeCun2015DeepLearning, LeCun1989}. 
The smaller network sizes required for success in decoding compared to visual tasks allows for state-of-the-art performance on limited amounts of data~\cite{Glaser2020}. 
However, these deep learning applications to neural decoding utilize architectures that ignore the underlying structure of the input neural activity.

One approach is to look at the underlying graphical structure of the neurons and the neuronal maps and utilize this information for feature extraction.
Graph Neural Networks (GNNs) feed in 1-dimensional connectivity information into a neural network and use that information to update the neural weights \cite{bessadok2023gnn}.
Although graphs are able to capture pairwise connectivity, neurons in the brain form dense connections that lead to heavily correlated activity across multiple neurons. 
Beyond these structural connections, higher-dimensional functional connectivity has been observed within groups of neurons exhibiting similar firing properties; for example, grid cells within a module~\cite{Hafting2005}.
Simplicial complexes, topological spaces with the ability to describe multi-way relationships, naturally lend themselves to defining and encapsulating the hierarchical properties of neuronal data~\cite{Hafting2005, Okeefe1976}, making them an increasingly popular tool for representing neural activity~\cite{Gardner2022, Curto2008, Giusti2015, Andjelkovic2020connectomes, Chaudhuri2019, Billings2021}. 
Hence, there exist simplicial convolutional neural networks (SCNNs) that account for this higher order connectivity~\cite{Ebli2020, Yang2022}. 

Our proposed approach, the simplicial convolutional recurrent neural network (SCRNN), combines the connectivity-based structure of the SCNN and the power of a RNN. 
First, the neural activity is defined on a simplicial complex via a preprocessing procedure.
Neural spikes are of binned to generate a binarized spike count matrix where each set of active cells within a time bin are connected by a simplex. 
The construction of the simplicial complex makes no assumptions about the spike train's encoding, and the higher dimensional connectivity of the simplicial complex ameliorates feature representation.
Then, each simplicial complex is fed into simplicial convolutional layers
% ~\cite{Hajij2021, Yang2022, Ebli2020, Bodnar2021} 
for feature extraction.
Next, the outputs of the final SC layer are concatenated to form a single feature vector which is fed into the RNN portion of the network.
Finally, the algorithm predicts either a head direction or a location, depending on the dataset used for training.
For an overarching view, see Figure~\ref{fig:intro}.

We first demonstrate the method by decoding head direction from a population of HD cells~\cite{Peyrache2015}, and compare the results to those produced by three other neural network (NN) architectures.
Applying the SCRNN to the head direction data provides the lowest average absolute error and mean absolute error compared to the three traditional neural networks we tested it against.
After verifying our architecture's viability on head direction decoding, we demonstrate the effectiveness of the SCRNN by decoding two-dimensional location from a population of grid cells and comparing it to the same networks mentioned above. 
We show that the SCRNN has the smallest average Euclidean distance between the ground truth and decoded location, demonstrating its aptitude for decoding different kinds of spiking data.
Notably, to the best of our knowledge, our grid cell decoding task marks one of the first deep learning applications to decoding experimental grid cell data. 

The paper is organized as follows. 
Section~\ref{sec:RW} examines related work and survey other relevant decoding algorithms. 
Section~\ref{sec:Methods} discusses the architecture of the SCRNN including the preprocessing procedure and the datasets we consider. 
Decoding results and comparisons to other machine learning algorithms can be found in Section~\ref{sec:Results} for both the head direction and grid cell data. 
Finally, we conclude and comment on future direction in Section~\ref{sec:Conclusion}.

%%%%%%%%%%%%%%%%%%%%%%%%%%%%%%%%%%%%%%%%%%%%%%%%
\section{Related Work}
\label{sec:RW}

The simplicial convolutional recurrent neural network (SCRNN) draws inspiration from the simplicial convolutional layer's ability to leverage the underlying connectivity of a dataset and the success of RNN's at decoding time-dependant data. 
First, we will look at how simplicial complexes have been used to capture connectivity.
Next, we will consider simplicial convolutional neural networks and how those have leveraged the underlying data structure for predictive purposes.
Finally, we consider prior instances decoding neural data, with an emphasis on machine learning methods.

%%%%%%%%%%%%%%%%%%%%%%%%%%%%%%%%
\subsection{Neural Decoding.} 
\label{subsec:nd}

Deep learning for neural decoding is a rapidly growing field due to neural networks' observed success at tasks like image recognition and sequence prediction and neural networks' ability to generalize beyond training data~\cite{Livezey2021deeplearing}. 
Neural networks have outperformed statistical methods at decoding head direction and two-dimensional position within an environment from neural recordings of HD cells and place cells, respectively~\cite{Xu2019, Frey2019, Tampuu2019}. 
The superior performance has been observed for a variety of network architectures including recurrent, fully-connected feed forward, and convolutional neural networks. 
The smaller network sizes required for success in decoding compared to visual tasks allows for state of the art performance on limited amounts of data~\cite{Glaser2020}.

%%%%%%%%%%%%%%%%%%%%%%%%%%%%%%%%
\subsection{Simplicial complexes and neural activity.}
\label{subsec:sc_neural_activity}

Simplicial complexes have previously been used to represent neural activity. 
The study in~\cite{Curto2008} used place cell spike trains to reconstruct the environment. 
The work in ~\cite{Giusti2015} analyzed clique complexes generated from place cell firing fields to detect geometric structure in matrices. 
Simplicial complexes also play a pivotal role in manifold discovery, a growing area of neuroscience focused on finding the underlying manifolds on which different types of neural activity live. 
Persistent homology on a point cloud representing the population activity of HD cells revealed the states of the HD circuit form a one-dimensional ring~\cite{Chaudhuri2019}. 
Similarly, persistent cohomology was employed to show that the activity of a single grid cell module forms a toroidal manifold~\cite{Gardner2022}.
For more background on simplicial complexes, see Subsection~\ref{subsec:SClayers}.

%%%%%%%%%%%%%%%%%%%%%%%%%%%%%%%%
\subsection{Simplicial convolutional neural nets}
\label{subsec:scnn}

Neural activity is regularly converted to a matrix where rows represent either individual neurons or different electroencephalogram (EEG) channels and columns correspond to non-intersecting time bins. 
The most common deep learning approach to handling the matrix is to use a convolutional neural network (CNN)~\cite{LeCun1989}. 
In a CNN, convolutional layers extract features from the input by aggregating weighted information from neighboring elements in the input matrix. 
This localization of information sharing assumes regular connectivity where only neighboring rows, or columns, bare significance to each other. 
But, the ordering of the matrix rows are arbitrary and not dependent on neural connectivity.
Hence, there is need for a different kind of convolution that takes into account the firing connectivity.

Simplicial convolutional neural networks (SCNNs), such as those found in \cite{Yang2022, Ebli2020}, which utilize the simplical complexes formed by the connectivity of the network as the input. 
These SCNNs take in the simplicial complexes constructed from the data and generate matrices that capture low and high dimensional connectivity information. 
These matrices are used to construct simplicial filters, which contain the neural networks weights. 
Then, these features are flattened and fed into a FFNN, which can use the features for prediction.
For our particular application, the prediction is a mouse's head direction or location. 
Although the simplical layers capture connectivity, spiking data's time-dependent nature makes other networks, such as RNNs, a better tool for neural decoding applications. 
As such, we propose a network that consists of simplicial convolutional layers and recurrent neural network layers. 

%%%%%%%%%%%%%%%%%%%%%%%%%%%%%%%%%%%%%%%%%%%%%%%%
\section{Methods}
\label{sec:Methods}
%%%%%%%%%%%%%%%%%%%%%%%%%%%%%
Our method consists of three major parts, preprocessing, SC layers, and the back end RNN. 
Below, we elaborate on each portion individually. 
For an overarching view of the architecture, see Figure~\ref{fig:intro}.

\begin{figure*}[h!]
    \centering
    \includegraphics[width=0.95\textwidth]{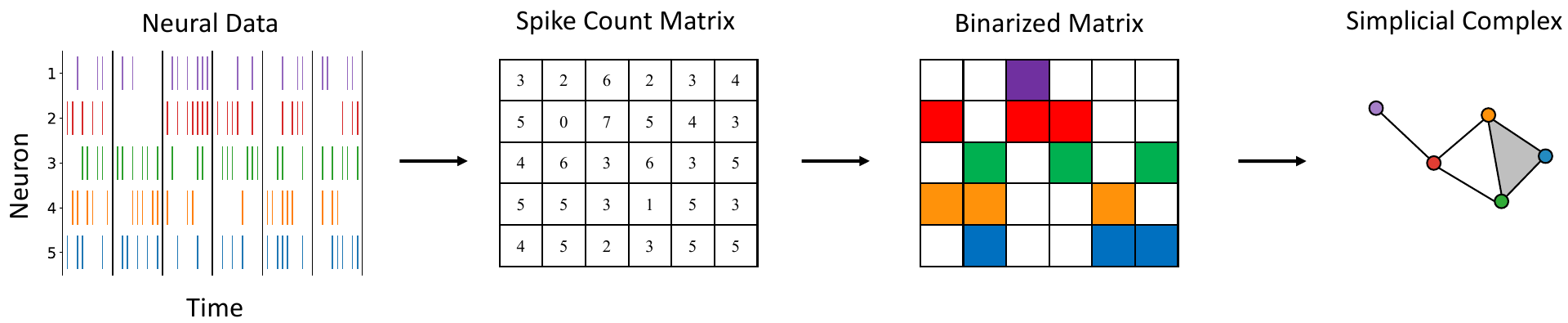}
    \caption{An example of the preprocessing procedure. 
    First, neural spiking data is represented as a raster plot. 
    Next, the data is binned and converted to a spike count matrix. 
    A row-wise thresholding procedure, given in Equation \eqref{eq:RowThresh}, binarizes the matrix. 
    In this figure, each colored box denotes a 1 and each white box denotes a 0. 
    Then, each neuron is represented by a node of the simplicial complex, color coded to match the corresponding matrix row.
    To construct the simplicial complex, the colored nodes are connected by the appropriate dimensional simplex to capture the neurons that fire together. 
    For example, we see that the second column of the binarized matrix has three active neurons (green, orange, and blue). 
    This generates a 2-simplex on the corresponding nodes. }
    \label{fig:pp}
\end{figure*}

%%%%%%%%%%%%%%%%%%%%%%%%%%%%
\subsection{Preprocessing}
\label{subsec:preprocessing}

One of the strengths of the SCRNN is its ability to process different types of data that do not have an explicit graphical structure. 
Neuronal spiking data is an example of data where extracting connectivity provides implicit structural information. 
Spiking data is captured by inserting probes into the brain and recording the electrical activity, specifically, when individual neurons fire. 
This data is captured in raster plots, where the x-axis represents time and the y-axis represents which neuron is firing. 
Hence, preprocessing spiking data into a simplicial complex provides information about which neurons fire together.

The experimental HD data and grid cell data consist of neurons and their corresponding spike times.
Given the spike times of $N$ simultaneously recorded neurons, we first construct a spike count matrix $A$ by creating $N_{b}$ non-intersecting bins of width $t_{bin}$ and counting each individual neuron's number of spikes within each bin, shown in Figure~\ref{fig:pp}. 
The element $A_{ij}$ is then set equal to the spike count of neuron $i$ within bin $j$. 
The next step is to binarize $A$ via a row-wise thresholding procedure. 
For a fixed row, consider the elements $\{ a_\ell \}^{N_{b}}_{\ell=1}$ ordered from highest to lowest. 
Then for some value $p\in(0, 1]$, we select $\{ a_\ell \}^{m^\star}_{\ell=1}$ for $m^\star$ given by,
\begin{eqnarray}
m^\star = \argmin_{1\leq m\leq N_{b}}\Bigg\{ \sum^{m}_{\ell=1} a_\ell  \ \ \  \mathrel{\Big|} \ \ \ \sum^{m}_{\ell=1} a_\ell\ \ \geq \ \ p \cdot\hspace{-0.5em} \sum^{N_{b}}_{\ell=1} a_\ell \Bigg\} \ \ . 
\label{eq:RowThresh}
\end{eqnarray}
The $m^\star$ selected row elements are then set to 1 while the remaining $N_{b} - m^\star$ elements are set to 0. 
This is repeated for every row of $A$ using the same value for $p$ as before. 
Note that thresholding row-wise accounts for the variability in total spikes among neurons by comparing each neuron's activity against itself. 
We then proceed column-wise through the binarized matrix, connecting each active neuron within a time bin by the appropriate-dimensional simplex, see Figure~\ref{fig:pp}. 
Specifically, if there are $0\leq n_{act}\leq N$ active neurons in a column, an $(n_{act} - 1)-$simplex is constructed on the nodes corresponding to those $n_{act}$ active neurons. 
This allows for a multi-way description of a group of nodes' relationship as opposed to the clique of 1-simplices that can only describe these nodes by their pairwise relationships.

%%%%%%%%%%%%%%%%%%%%%%%%%%%%%
\subsection{Simplicial Convolutional Layers}
\label{subsec:SClayers}

It is common practice for neural activity to be converted to a matrix where rows represent individual neurons and columns correspond to time bins. 
The most widely used deep learning approach to handling matrices as inputs is to employ a convolutional neural network (CNN). 
In a CNN, convolutional layers extract features from the input by aggregating weighted information from neighboring elements in the input matrix. 
This localization of information-sharing assumes regular connectivity where only neighboring rows, or columns, possess significance to each other.
Thus, in tasks where rows of a matrix neighboring each other bares no significance, CNNs do not intuitively extract features.

Simplicial convolutions generalize convolutions to account for data with irregular connectivity~\cite{Hajij2021, Yang2022, Ebli2020, Bodnar2021}. 
We introduce simplices and simplicial complexes, the topological structures we exploit for feature representation. 
For more information on simplicial complexes beyond what is outlined below, see~\cite{Hatcher2002}.

\begin{figure*}[hbt!]
    \centering
    \includegraphics[width=\textwidth]{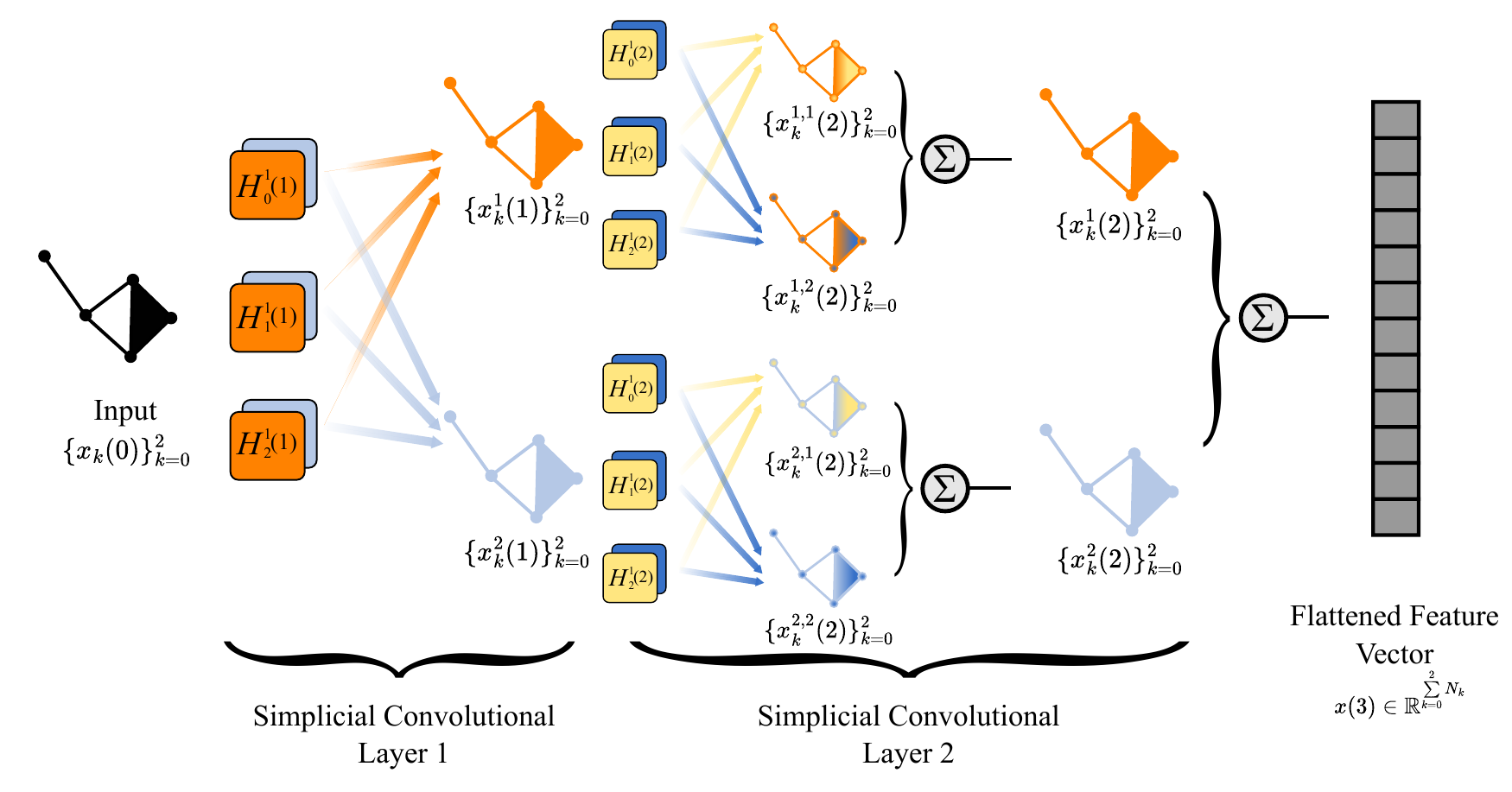}
    \caption{
    A diagram of $L=2$ simplicial convolutional layers each equipped with two filters, $H^1_k (\_)$ and $H^2_k (\_)$, for each simplicial dimension $k=0, 1, 2$, and $F=2$ filters for each dimension of the input simplicial complex. 
    In the first layer, we see three orange and three blue filters indicating the three dimensions of the input simplicial complex. 
    The features extracted using these filters result in new orange and blue simplicial complexes, respectively. 
    In the second simplicial convolutional layer, the process is repeated with two new filters, depicted by yellow and dark blue, giving us 4 simplicial complexes. 
    In order to prevent exponential growth, features extracted from the same input from the previous layer are summed, resulting in a new orange and blue simplicial complex. 
    Finally, all extracted features are summed and flattened to create one feature vector for the RNN.}
    \label{fig:sc_layers}
\end{figure*}

\begin{definition} A collection $\{ v_0, v_1, \dots, v_n \}\subset\RR^d\setminus \{ 0 \}$ is {\it geometrically independent} if and only if for any $\{ t_0, t_1, \dots, t_n \}\subset\RR$ with $\sum^n_{i=0} t_i = 0$, the condition $\sum^n_{i=0} t_i v_i = 0$ implies $t_i = 0$ for all $i\in\{0, 1, \dots, n \}$. 
\end{definition}
\begin{definition} A {\it $k-$simplex}, $s^k$, is the convex hull of $k+1$ geometrically independent points $\{ v_0, v_1, \dots, v_k \}$, denoted by $\left[ v_0, v_1, \dots, v_k \right]$. 
\end{definition}
\begin{definition} The {\it faces} of a $k-$simplex $\left[ v_0, v_1, \dots, v_k \right]$ are the $(k - 1)-$simplices given by $\left[ v_0, \dots, v_{j-1}, v_{j+1}, \dots, v_k \right]$ for some $j\in\{ 0, 1, \dots, k \}$ and are denoted $s^{k-1}_j \subset s^k$. 
\end{definition}
\begin{definition} A {\it simplicial complex} $S$ is a collection of simplices satisfying 
\begin{enumerate}
	\item if $s\in S$, then every face of $s$ is in $S$ and
	\item if $s_1, s_2 \in S$, then $s_1 \cap s_2 = \emptyset$ or $s_1 \cap s_2 \in S$ \ .
\end{enumerate}
\end{definition}
To ease understanding, one may consider a 0-simplex as a vertex, a 1-simplex as an edge, a 2-simplex as a triangle, a 3-simplex as a tetrahedron, and so on. Orientation can be assigned to $k-$simplices forming what is called an {\it ordered $k-$simplex}. 
For a face $s^{k-1} \subset s^k$, if the orientation of $s^{k-1}$ coincides with that of $s^k$, we write $s^{k-1} \prec s^k$. 
Additionally, features, typically vectors or scalars, can also be assigned to the simplices. 
The features of the $k-$simplices are represented by a vector, or matrix depending on the feature size, called the {\it $k-$cochain}, and it is denoted by $c_k$.
\begin{definition} Let $\{ s^k_i \}^{N_k}_{i=1}$ be the ordered $k-$simplices of a simplicial complex. Then for each $s^k_i \in \{ s^k_i \}^{N_k}_{i=1}$, assign a feature $c^i\in\RR^{N_\text{feat}}$. The \emph{$k-$cochain}, $c_k \in\RR^{N_k \times N_{\text{feat}}}$, is then given,
\begin{eqnarray}
\left[ c_k \right]_{ij} = \left[ c^i \right]_j \ .
\label{eq:cochain}
\end{eqnarray}
\end{definition}

For these layers, input data is defined on a simplicial complex, and information-sharing is generated by the {\it Hodge-Laplacian}. 
To define the Hodge-Laplacian, we must first introduce the {\it $k-$dimensional incidence matrix}, $B_k\in\RR^{N_{k-1} \times N_k}$, where the $ij$th element is given by,
\begin{eqnarray}
\left[ B_k \right]_{ij} = 
\begin{cases}
	0, &\text{if } s^{k-1}_i \not\subset s^k_j, \\
	-1, &\text{if } s^{k-1}_i \subset s^k_j \text{ and } s^{k-1}_i \not\prec s^k_j \ \ \ ,\\
	1, &\text{if } s^{k-1}_i \subset s^k_j \text{ and } s^{k-1}_i \prec s^k_j
\end{cases}
\label{eq:IncMat}
\end{eqnarray}
where $N_{k-1}$ and $N_k$ are the number of $(k-1)-$simplices and $k-$simplices, respectively. 
Note, we consider $B_0~=~0~\in~\RR^{N_0 \times N_0}$. 
Then, finally, the $k-$Hodge-Laplacian, $L_k\in\RR^{N_k\times N_k}$, is defined as,
\begin{eqnarray}
L_k = B_k^T B_k + B_{k+1} B_{k+1}^T \ . 
\label{eq:HodgeLap}
\end{eqnarray}

In simplicial convolutions, the terms of the Hodge-Laplacian in Equation (\ref{eq:HodgeLap}) act as shift-operators defining which simplices of the same dimension share information. 
The terms $B_k^T B_k$ and $B_{k+1} B_{k+1}^T$ are called the {\it lower} and {\it upper Laplacian}, and they capture connectivity by lower and higher dimensional simplices, respectively.
A {\it degree D simplicial filter} consisting of weights $W~=~\{ W_i\}_{i=0}^{2D}$ is an operator, $H_k\in\RR^{N_k\times N_k}$, given by,
\begin{eqnarray}
H_k = W_0 I + \sum_{i=1}^{D} W_i (B_k^T B_k)^i +  \sum_{i=1}^{D} W_{i + D} ( B_{k+1} B_{k+1}^T)^i \ , 
\label{eq:SimpFilt}
\end{eqnarray}
where $k$ is the dimension of simplices and $(\cdot)^i$ denotes the $i-$th power of a matrix. 
Note, each power of the lower and upper Laplacians localizes information-sharing to within the $i$ nearest $k-$simplices, similar to increasing the filter size in a traditional convolutional layer.

We now discuss the dynamics of the \emph{simplicial convolutional layers} of an SCRNN. The proof of the following proposition is delegated to the Supplementary Materials.

\begin{table*}[ht!]
    \centering
\begin{tabular}{| l || p{3.5em} | p{3.5em} | p{3.5em} | p{3.5em} || p{3.5em} | p{3.5em} | p{3.5em} | p{3.5em} |}
        \hline
        & \multicolumn{4}{|c||}{Head Direction Decoding Lowest AAE} & \multicolumn{4}{|c|}{Grid Cell Decoding Lowest AED}\\ \hline
        \hline
        Hyperparameters     & FFNN  & RNN   & GNN   & SCRNN & FFNN  & RNN   & GNN   & SCRNN \\
        \hline\hline
        Epochs              & 100   & 50    & 100   & 100   & 100   & 100   & 100   & 100   \\ \hline
        Batch Size          & 32    & 16    & 64    & 8     & 32    & 32    & 8     & 8     \\ \hline
        Learning Rate       & 0.001 & 0.0001& 0.001 & 0.0001& 0.001 & 0.001 & 0.001 & 0.001 \\ \hline
        Dropout             & 0.2   & 0.2   & 0.3   & 0.3   & 0.2   & 0.3   & 0.2   & 0.2   \\ \hline
        NN Layers           & 2     & 2     & 2     & 2     & 3     & 3     & 1     & 1     \\ \hline
        Layer Width/Hidden Size& 128& 200   & 100   & 200   & 512      & 200   & 100   & 50    \\ \hline
        SC Layers           &       &       & 1     & 2     &       &       & 2     & 1     \\ \hline
        \# of Filters       &       &       & 3     & 2     &       &       & 3     & 3     \\ \hline
        Seq.~Length         &       &       & 5     & 5     &       &       & 5     & 5     \\ \hline
    \hline
        Validation loss     & 0.380 & 0.221 & 0.418 & 0.321 & 0.015 & 0.001 & 0.003 & 0.002 \\ \hline
        Training MAE (HD)   & 10.959& 9.414 & 9.406 & 6.785 &       &       &       &       \\ \hline 
        Training AAE (HD)   & 15.918& 12.548& 12.804& 8.233 &       &       &       &       \\ \hline
        Test MAE (HD)       & 12.080& 9.812 & 9.950 & 8.416 &       &       &       &       \\ \hline 
        Test AAE (HD)       & 17.990& 14.587& 14.624& 11.493&       &       &       &       \\ \hline
        Training AED (grid) &       &       &       &       & 7.620 & 3.086 & 3.030 & 2.547 \\ \hline
        Validation AED (grid) &     &       &       &       & 11.801& 3.350 & 3.539 & 3.088 \\ \hline
\end{tabular}
    \caption{A table comparing the different networks on the head direction data and the grid cell data based on their trial with the lowest AAE, measured in degrees and AED, measured in centimeters, respectively. 
    Note, blank cells indicate that the hyperparameter was not used for that network.}
    \label{tab:comparison}
\end{table*}

\begin{proposition} Consider an SCRNN consisting of $L$ simplicial convolutional layers, each equipped with $F$ filters, $\{ H^f_k(\ell) \}^F_{f=1} $, for each dimension $k$ of the functional simplicial complex with maximum simplicial dimension $K$, where $\ell\in\{1, 2, \dots, L\}$ denotes the simplicial convolutional layer.
In such a network, the number of parameters used in the simplicial convolutional layers is $F [2(D+1) + (K-1)(2D+1)] L$. 
\end{proposition}
Note that the dynamics of the simplicial convolutional layers prevent exponential growth of parameters with respect to filters and number of layers. 

For the first layer $\ell=1$, features $\{ \mathbf{x}^f_k(1) \}^F_{f=1}$ are extracted from the input, $\mathbf{x}_k(0)$, via some nonlinear transformation $\sigma$,
\begin{eqnarray}
\mathbf{x}^f_k(1) = \sigma \left( H^f_k(1) \ \mathbf{x}_k(0) \right) \ , 
\label{eq:SimpConv1}
\end{eqnarray}
for each $f=1, 2, \dots, F$ and $k=1, 2, \dots, K$. 
Note $\mathbf{x}_0(0)\in\RR^{N_0\times n_{col}}$ for some hyperparameter $1 \leq n_{col} \leq N_{b}$, and $\mathbf{x}_k(0)\in\RR^{N_k}$ for $1\leq k \leq K$.
For the intermediate simplicial convolutional layers $\ell = 2, 3, \dots, L - 1$ and fixed $k$, each of the filters $\{ H^f_k(\ell) \}$ is applied to each of the extracted features from the previous layer. 
To prevent the exponential growth of the number of filters, the outputs extracted from the same feature from the previous layer are summed together to create one single output feature. 
That is, for each feature $\{ \mathbf{x}^g_k(\ell - 1) \}^F_{g=1}$ from the previous layer, we extract,
\begin{eqnarray}
\mathbf{x}^g_k(\ell) = \sigma \left( \sum^F_{f=1} H^f_k(\ell) \ \mathbf{x}^g_k(\ell - 1) \right) \ , 
\label{eq:SimpConvInter}
\end{eqnarray}
for $g\in\{1, 2, \dots, F\}$.
In the final simplicial convolutional layer, $\ell=L$, features are extracted following the same procedure as the intermediate layers, but additionally, all extracted features are summed:
\begin{eqnarray}
\mathbf{x}_k(L) = \sum^F_{g=1} \mathbf{x}^g_k(L) = \sum^F_{g=1} \sigma \left( \sum^F_{f=1} H^f_k(\ell) \ \mathbf{x}^g_k(\ell - 1) \right) \ ,
\label{eq:SimpConvFinal}
\end{eqnarray}
where $\mathbf{x}_k^g(L)$ as in Equation \eqref{eq:SimpConvInter} for $\ell = L$.
If $1 < n_{col}$, then $\mathbf{x}_0(L)$ is summed across columns, which gives us $\mathbf{x}_0(L)\in\RR^{N_0}$.
Finally, the outputs for each dimension of the simplicial complex, $\{ \mathbf{x}_k(L) \}^K_{k=1}$, are stacked to create one output feature vector, $\mathbf{x}(L)\in\RR^{\sum^K_{k=0}N_k}$. 
For illustrative purposes, Figure~\ref{fig:sc_layers}, depicts $L=2$ simplicial convolutional layers each consisting of $F=2$ filters for each dimension of the input simplicial complex. 

%%%%%%%%%%%%%%%%%%%%%%%%%%%%%%%%%%%
\subsection{Simplicial Convolutional Recurrent Neural Network}
\label{subsec:SCRNN}
To form an input sequence to the RNN component of the SCRNN, we consider the outputs of the simplicial convolutional layers corresponding to a desired number of consecutive time bins.
Given the sequential nature of the decoding task, we append the simplicial convolutional layers with a multi-layer RNN, a neural network architecture designed for time-series data.
We opt for the Elman RNN architecture~\cite{Elman1990} over its more complex counterpart, the long short-term memory network (LSTM)~\cite{Hochreiter1997}. 
In this task, only neural activity recorded in time bins close to the target time bin bare any relevance to the decoded variable, thus, making the extra parameters in an LSTM designed for handling long sequences unnecessary.
Elman RNNs utilize what are called hidden states to handle sequential data.
Specifically, for a given input sequence ${ \mathbf{z}_{t} }$, an Elman RNN computes hidden state, $h_t$, given by,
\begin{eqnarray}
h_t = \sigma \left( W_h \mathbf{z}_{t} + b_h + W_c h_{t - 1} + b_c \right) \ , 
\label{eq:RNNhidden}
\end{eqnarray}
where $W_h, W_c$ are weight matrices and $b_h, b_c$ are bias vectors.
The final output of an RNN is obtained by computing the nonlinear mapping of a linear transformation of the hidden state; that is,
\begin{eqnarray}
y_t = \sigma \left( W h_t + b \right) \ ,
\label{eq:RNNout}
\end{eqnarray}
where $h_t$ as in Equation \eqref{eq:RNNhidden}. 
Finally, a multi-layer RNN is created by stacking multiple RNNs, feeding the outputs of one as the inputs to another.
In the SCRNN, simplicial complexes generated from consecutive time bins are fed as inputs to the simplicial convolutional layers, and their outputs form the input sequence to the backend RNN.

%%%%%%%%%%%%%%%%%%%%%%%%%%%%%%%%%%%%%%%%%%%%%%%%
\section{Results}
\label{sec:Results}

For each task, we compare four different networks; the FFNN, RNN, GNN, and SCRNN. 
We optimize the hyperparameters of each network using RayTune, a distributed hyperparameter tuning tool \cite{liaw2018tune}.
The program uses the Tree-structured Parzen Estimators (TPE), an algorithm that combines random search with two greedy sequential methods~\cite{bergstra2011algorithms}. 
Head direction decoding accuracy was measured two different ways. 
First, we considered the median absolute error (MAE), which is defined, 

\begin{eqnarray}
MAE = \mae_{n = 1,\ 2,\ ...,\ N_{b}} \left| \ \textit{rescale}\big[ \theta_{dec}(n) - \theta_{true}(n) \big] \ \right| \ ,
\label{eq:MAE}
\end{eqnarray}
where $N_{b}$ is the number of time bins, and $\theta_{dec}, \; \theta_{true}\in[0^\circ,360^\circ)$ are the decoded and the ground truth directions, respectively. 
The mapping \textit{rescale} accounts for the ring structure of HD. 
For example, $310^{\circ}$ and $20^{\circ}$ should be recorded as a difference of $70^{\circ}$ instead of $290^{\circ}$. 

Similarly, we compute the average absolute error (AAE), which considers the average instead of the median discrepancy as defined below, 

\begin{equation}
    AAE = \frac{1}{N_{b}} \sum_{n=1}^{N_{b}}\left| \ \textit{rescale}\big[ \theta_{dec}(n) - \theta_{true}(n) \big] \ \right| \ .
\label{eq:AAE}
\end{equation}

While optimizing our head direction networks, we chose to minimize AAE. 
The MAE is included for additional comparison.

To measure the success of our grid cell model, we compute the Average Euclidean Distance (AED) across all time-bins:
\begin{eqnarray}
AED = \frac{1}{N_{b}} \sum^{N_{b}}_{n=1} \sqrt{\left( x_{d}(n) - x_{t}(n) \right)^2 + \left( y_{d}(n) - y_{t}(n) \right)^2 } \ \ ,
\label{eq:AED}
\end{eqnarray}
where $N_{b}$ is the number of time bins, and $\left(x_{d}, y_{d}\right),$ $\left(x_{t}, y_{t}\right)$ are the decoded, and the ground-truth $xy-$coordinates, respectively.

To evaluate our architecture, we look at two different types of spiking data, head direction spiking data and grid cell spiking data, both of which are outlined in detail below.

\begin{figure}
\begin{tabular}{c}
\textbf{a) Method: FFNN \hspace{3em} Test AAE: 12.738}\\
    \includegraphics[width = 0.43\textwidth]{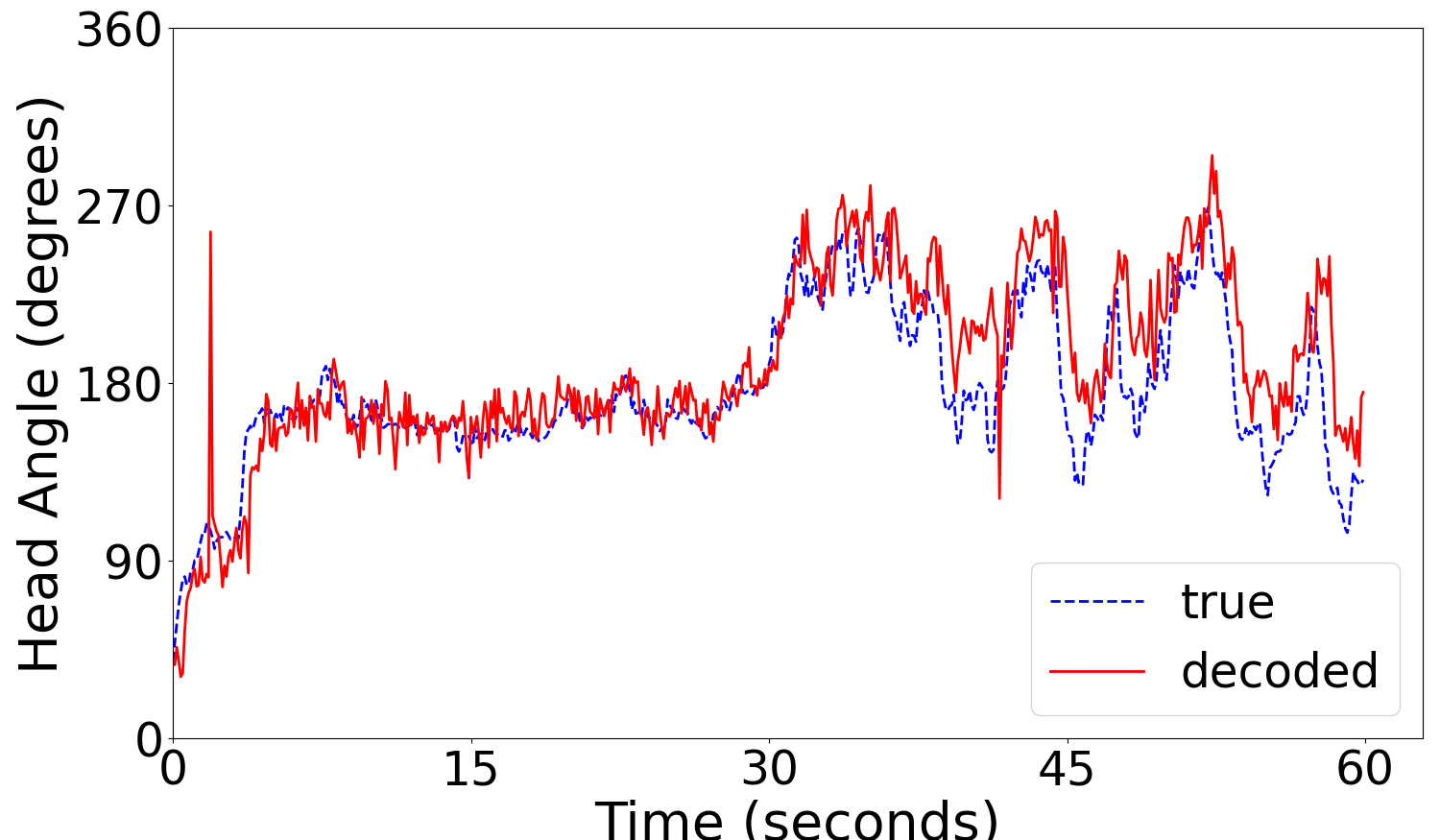}\\\vspace{0.5em}
\textbf{b) Method: RNN \hspace{3em} Test AAE: 8.636}\\
    \includegraphics[width = 0.43\textwidth]{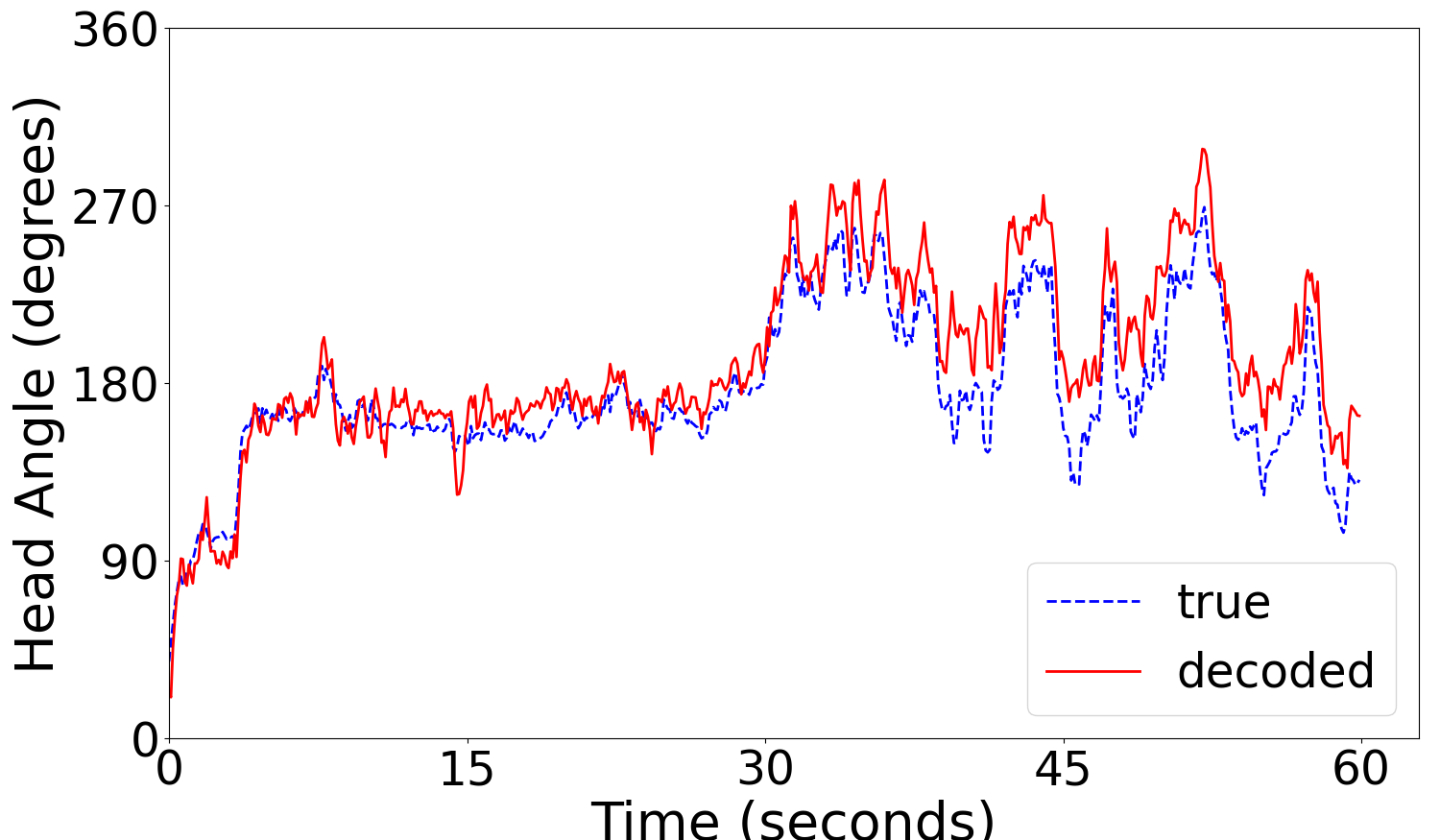} \\\vspace{0.5em}
\textbf{c) Method: GNN \hspace{3em} Test AAE: 8.584}\\
    \includegraphics[width = 0.43\textwidth]{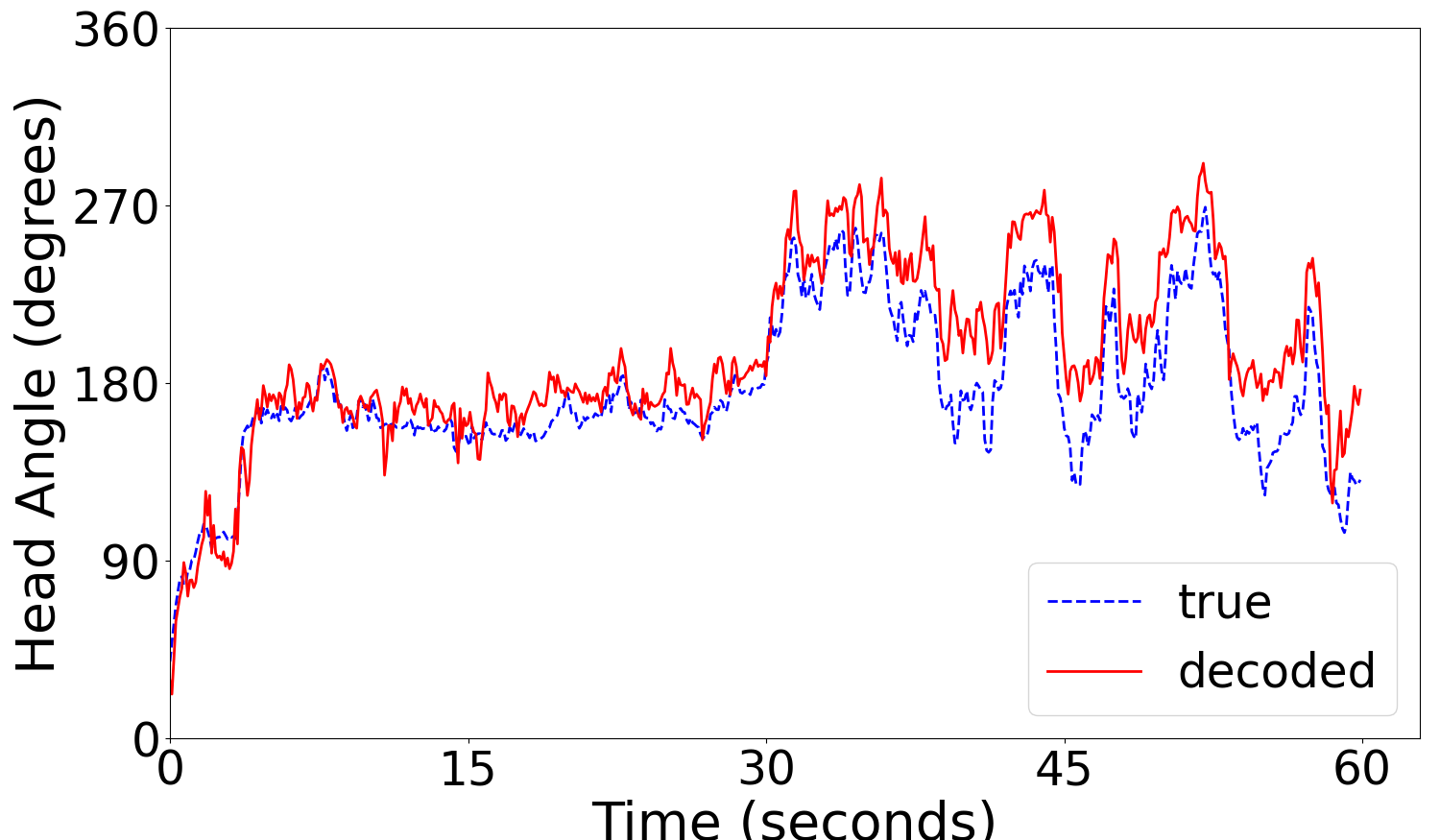}\\\vspace{0.5em}
\textbf{d) Method: SCRNN \hspace{3em} Test AAE: 7.407}\\
    \includegraphics[width = 0.43\textwidth]{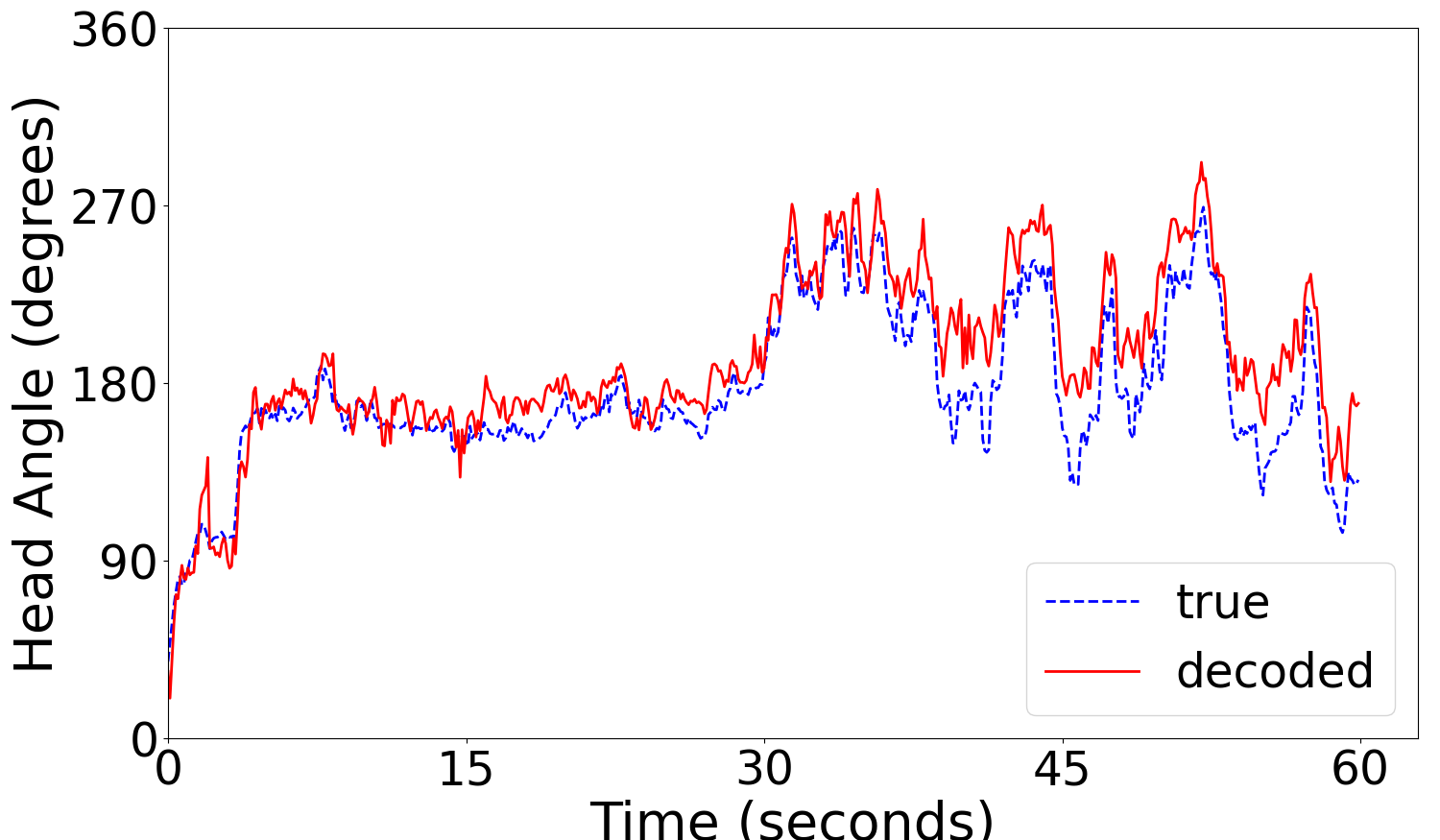} 
    \end{tabular}
    \caption{Plots depicting the true and the predicted head angles for the first minute of four different networks, a) FFNN, b) RNN, c) GNN, and d) SCRNN and the corresponding Test AAEs for the first 5 minutes. Each network was generated with the RayTune optimized hyperparameters listed in Table~\ref{tab:comparison}.}
    \label{fig:HD_plots}
\end{figure}

%%%%%%%%%%%%%%%%%%%%%%%%%%%%%%%%%%%%%%%%%%%%%%%%
\subsection{Head Direction}
\label{subsec:hd}

The neurons making up the head direction (HD) system in the brain encode the direction the head is facing at any given time.
This encoding is done by identifying different ensembles of certain neurons, called HD cells, which fire simultaneously, where each grouping of cells represents a different direction. 
Additionally, HD is decoded independently of body orientation.

\begin{figure*}[ht]
    \centering
    \includegraphics[width = 0.98\textwidth]{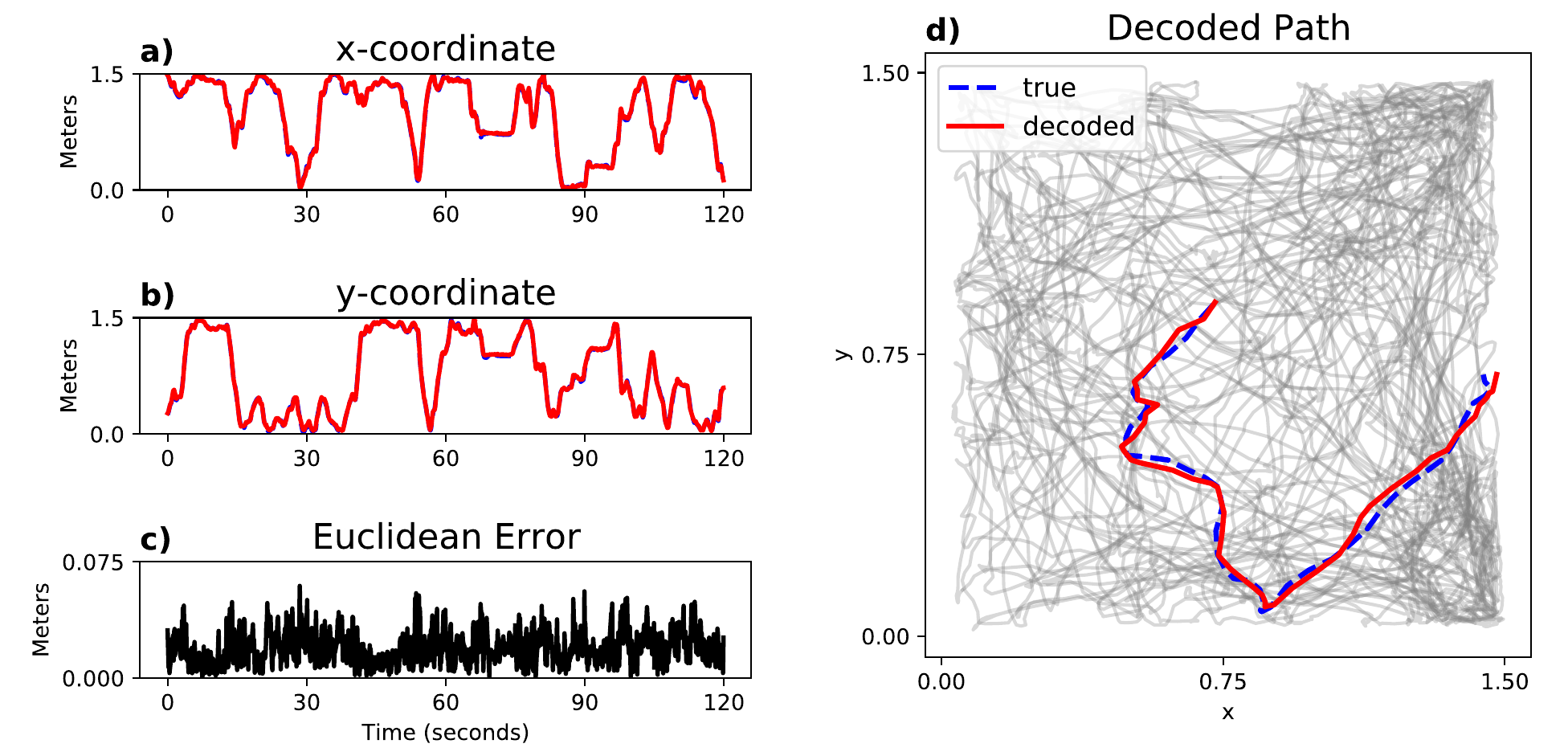}
    \caption{a)-d) Plots showing results from two minutes of the grid cell decoding task. 
    a)-b) Comparison of decoded vs.~ground truth x$-$coordinate and y$-$coordinate. 
    c) Error for each time bin measured by Equation \eqref{eq:AED}. 
    d) In grey, the ground truth position of the rat's trajectory in the environment. 
    For visual purposes, we include colorized paths showing a 5 second comparison of decoded vs.~ground truth position. }
    \label{fig:gc}
\end{figure*}

To demonstrate the effectiveness of our method, we analyze HD data recorded in \cite{Peyrache2015}.
The spike times of HD cells in the anterodorsal thalamic nucleus (ADn) along with the corresponding ground truth head angles of 7 mice were recorded using multi-site silicon probes and an alignment of LED lights on the mice's headstage, respectively. 
The sessions recorded comprised of 2 hours of sleep followed by 30-45 minutes of foraging in an open rectangular environment followed by 2 more hours of sleep. 
For the following analysis, we used the foraging portion of session `Mouse28-140313', which consists of recordings from 22 head direction cells.

We used RayTune to search the parameter space given in Table~\ref{tab:hd_hyper_parameters} in the Supplementary Material to optimize the AAE for each of the four networks.
The FFNN recorded $\text{AAE}=17.990^\circ$, the RNN yielded $\text{AAE}=14.587^\circ$, the GNN $\text{AAE}=14.624^\circ$, whereas the SCRNN produced $\text{AAE}=11.493^\circ$.
For the corresponding hyperparameters and the MAE, please see Table~\ref{tab:comparison}.
All around the SCRNN performed the best across all different architectures, as shown in Figure~\ref{fig:HD_plots}.

%%%%%%%%%%%%%%%%%%%%%%%%%%%%%%%%%%%%%%%%%%%%%%%%
\subsection{Grid Cells}
\label{subsec:gc}

Grid cells encode two-dimensional allocentric location by forming hexagonal, periodic firing fields within an environment. 
Grid cells with firing fields exhibiting the same spacing and orientation form what are referred to as modules. 
Because firing fields for cells within a module are the same, except for a shift in space, it takes more than one module to encode position~\cite{Mathis2012, Stemmler2015}. 
Cells firing at the same time within a module generate a spatial grid over the environment. 
A multi-scale representation for location is then created by layering the grids generated by different modules. 

To showcase the ability of our method on a more complex task, we consider a population of cells recorded in layers II and III of the Medial Entorhinal Cortex (MEC) of a moving rat, which contains `pure' grid cells, HD cells, and conjunctive grid/HD cells~\cite{Gardner2022, Gerlei2022gchd}. 
Neural activity was recorded using high-site-count Neuropixels silicon probes while the rat foraged in an open square environment.
Specifically, we look at a population of 482 cells that contains three grid modules consisting of 166, 167, and 149 cells total with 93, 149, and 145 of them being pure grid cells, respectively. 
The rest of the population is made up of conjunctive grid/HD cells.
The difficulties of decoding such a population stem from not only the large amount of cells, but also the fact that some cells are not solely responsible for encoding position, the target variable we aim to decode~\cite{McNaughton2006}.
We decode the position (as in xy$-$coordinates) from the activity of a population of cells recorded in a moving rat, which contains pure grid cells and conjunctive grid/HD cells~\cite{Gardner2022}.
The larger population of cells and, consequently, the larger size of the functional simplicial complex compared to the HD decoding task, means a more heavily parameterized SCRNN is required to decode position.

We used RayTune to search the hyperparameter space to minimize AED (Equation~(\ref{eq:AED})), see Table~\ref{tab:hd_hyper_parameters}.
For the training set, the FFNN recorded $\text{AED}=7.620$ cm, the RNN yielded $\text{AED}=3.086$ cm, the GNN $\text{AED}=3.030$ cm, whereas the SCRNN produced $\text{AED}=2.547$ cm.
For the vaildation set, the FFNN recorded $\text{AED}=11.801$ cm, the RNN yielded $\text{AED}=3.350$ cm, the GNN $\text{AED}=3.539$ cm, whereas the SCRNN produced $\text{AED}=3.088$ cm.
Plots of the best predictions for the hyperparameters associated with each network type are shown in Figure~\ref{fig:gc}.
For the corresponding hyperparameters, please see Table~\ref{tab:comparison}.
Thus, the SCRNN is clearly able to learn the pattern between grid cell activity and position in the environment. 
Note, a discrepancy between training and test results is expected given the fact grid cells may not encode the exact location, so training could bias the network to map neural codes for general locations to the specific labelled locations included in the training data. 

%%%%%%%%%%%%%%%%%%%%%%%%%%%%%%%%%%%%%%%%%%%%%%%%
\section{Conclusion}
\label{sec:Conclusion}

As neuroscientists capture more data, they desire tools not only decode neural data, but also leverage the underlying structure of the brain's systems to predict animal behavior.
This additional requirement provides interpretability, something traditional neural networks lack. 
The SCRNN combines the interpretability of the simplicial convolutional layers with the power of a RNN. 
As shown on the head direction and grid cells, the SCRNN is able to decode spiking data with a higher level of accuracy than traditional neural network architectures. 
Indeed, the SCRNN outperformed the FFNN, RNN, and GNN when evaluated by the mean absolute error and the average absolute error for the head direction data and also had lowest average Euclidean distance of the three networks for the grid cell decoding task.
Future work includes applying the SCRNN to other decoding tasks, such as decoding place cells and cue cells. 
From a modeling stand point, it would be crucial to examine how higher dimensional simplices as expressed in highly correlated data may be used to capture more information, and decode even more of the underlying structure of brain activity.

%%%%%%%%%%%%%%%%%%%%%%%%%%%%%%%%%%%%%%%%%%%%%%%%
\paragraph{Code availability}
The code can be found on GitHub at \url{https://github.com/emitch27/SCRNN}.

%%%%%%%%%%%%%%%%%%%%%%%%%%%%%%%%%%%%%%%
\paragraph{Acknowledgment} This work has been partially funded by the US Army Research Lab Contract No W911NF2120186. 

%%%%%%%%%%%%%%%%%%%%%%%%%%%%%%%%%%%%%%%
\bibliography{ref}

%%%%%%%%%%%%%%%%%%%%%%%%%%%%%%%%%%%%%%%%%%%%%%%%
\onecolumn
\section*{Supplementary Material}
\label{sec:supp_material}

\subsection*{Hyperparameter tuning}
Below, we outline the different hyperparameters used throughout tuning for the results included in the main paper. 
For all hyperparameter tuning, we used RayTune, a python package that uses a Tree-structured Parzen Estimator approach to search the hyperparameter space. 
These hyperparameters and corresponding ranges are outlined in Table~\ref{tab:hd_hyper_parameters}.
Note, for each network, we used the ReLU activation function, a threshold of 30 and for the GNN and SCRNN, we used a sequence length of 5 when calculating the graph and simplicial complex, respectively.

\paragraph{HD decoding hyperparameters.}
The training and test data was constructed from 20 minutes of a 38 minute session of open foraging using $t_{bin} = 100$ms. 
The first 25\% of the data was used for testing data and the last 75\% of the data was used for training.
Ground truth labels were computed by taking the circular mean of recorded head directions within each time bin.
During construction of the functional simplicial complex, the maximum dimension of simplices was bounded at $k=2$.
This bound was chosen due to the computational cost associated to including higher dimensional complexes.
We then used RayTune to search the hyperparameter space to identify the best hyperparameters for each network, which can be found in \ref{tab:hd_hyper_parameters}. 

\paragraph{Grid cell decoding hyperparameters}
We use 10 total minutes of recorded neural activity and ground truth position with bin sizes of $t_{bin}=100$ ms.
The first 20\% of the data was used for testing data, and the last 80\% of the data was used for training.
Ground truth position is computed as an average of observed positions within a time bin.
The features extracted from the simplicial convolutional layers are then fed to a RNN with 3 blocks using a hidden dimension of size 50.
The network trained for 50 epochs on a batch size of 16 with learning rate 0.001.
Similar to the HD decoding task, framework hyperparameters were tuned with RayTune to minimize AED.

\begin{table*}[h!]
\centering
    \begin{tabular}{|c|c|c|c|}
        \hline
        \multicolumn{4}{|c|}{Head Direction Decoding Hyper Parameter Search}\\ \hline
        \hline
        & FFNN & RNN & GNN/SCRNN\\
        \hline\hline
        Epochs              & [25, 50, 100] & [25, 50, 100] &[50, 100]\\ \hline
        Batch Size          & [8, 16, 32] & [8, 16, 32, 64] &[8, 16, 32, 64]\\ \hline
        Learning Rate       & [0.01, 0.001, 0.0001] & [0.01, 0.001, 0.0001, 0.00001] & [0.001, 0.0001, 0.00001]\\ \hline
        Dropout             & [0.2, 0.3, 0.4] & [0.2, 0.3, 0.4] & [0.2, 0.3, 0.4]\\ \hline
        FFNN/RNN Layers     & [2, 3, 4] & [1, 2, 3] & [1, 2, 3]\\ \hline
        Layer Width         & [64, 128, 256] &  & [32, 64, 128] \\ \hline
        Hidden Size         && [50, 100, 200] & [50, 100, 200]\\ \hline
        Degree              &&& [1, 2]\\ \hline
        SC Layers           &&& [1, 2, 3, 4]\\ \hline
        \# of Filters       &&& [1, 3, 5]\\ \hline
        Degree              &&& [1, 2]\\ \hline
        \hline
        \multicolumn{4}{|c|}{Grid Cell Decoding Hyper Parameter Search}\\ \hline
        \hline
        & FFNN & RNN & GNN/SCRNN\\
        \hline\hline
        Epochs              & [50, 100] & [25, 50, 100] &[50, 100]\\ \hline
        Batch Size          & [8, 16, 32] & [8, 16, 32, 64]  & [8, 16]\\ \hline
        Learning Rate       & [0.001, 0.0001, 0.00001] & [0.001, 0.0001, 0.00001] & [0.001, 0.0001, 0.00001] \\ \hline
        Dropout             & [0.2, 0.3, 0.4] & [0.2, 0.3, 0.4, 0.5] & [0.2, 0.3, 0.4] \\ \hline
        FFNN/RNN Layers     & [2, 3, 4] & [1, 2, 3] & [1, 2, 3]\\ \hline
        Layer Width         & [128, 256, 512] & & [32, 64, 128, 256]\\ \hline
        Hidden Size         &  & [100, 200, 400] &[50, 100, 200]\\ \hline
        SC Layers           &&& [1, 2, 3]\\ \hline
        \# of Filters       &&& [1, 3, 5]\\ \hline
        Degree              &&& [1, 2]\\ \hline
        \end{tabular}
    \caption{Search values used for the RayTune hyperparameter tuning of the head direction and grid cell networks.  
    All hyperparameter searches were conducted with 15 minutes of training data and 5 minutes of testing data.
    Note, blank cells indicate that the hyperparameter was not used for that network.}
    \label{tab:hd_hyper_parameters}
\end{table*}

\subsection*{Proof of Proposition 1}
\label{sec:SuppPropProof}
Let $f\in\{ 1, 2, \dots, F \}$ and $\ell\in\{ 1, 2, \dots, L \}$ be arbitrary. Fix $k=0$. Then because $B_0~=~0~\in~\RR^{N_0 \times N_0}$, we have
\begin{eqnarray}
H^f_0(\ell) = W^{f, 0}_0(\ell) I + \sum_{i=1}^{D} W^{f, 0}_{i} (\ell) ( B_1 B_1^T)^i \ , 
% \label{eq:}
\end{eqnarray}
where $\{ W^{f, 0}_i (\ell) \}^{D}_{i=0}$ are filter parameters. 
Thus, the 0-dimensional component of an arbitrary filter in an arbitrary simplicial convolutional layer contains $D+1$ parameters.
Similarly, for fixed $k=K$, we have
\begin{eqnarray}
H^f_K(\ell) = W^{f, K}_0(\ell) I + \sum_{i=1}^{D} W^{f, K}_{i}(\ell) ( B_K^T B_K)^i \ . 
% \label{eq:}
\end{eqnarray}
Therefore, the $K$-dimensional component of an arbitrary filter in an arbitrary simplicial convolutional layer also contains $D+1$ parameters.
Now, for an intermediate dimension $k\in\{1, 2, \dots, K-1 \}$, a filter is defined
\begin{eqnarray}
H^f_k(\ell) = W^{f, k}_0(\ell) I + \sum_{i=1}^{D} W^{f, k}_{i}(\ell) ( B_k^T B_k)^i + \sum_{i=1}^{D} W^{f, k}_{i+D}(\ell) ( B_{k+1} B_{k+1}^T)^i \ , 
% \label{eq:}
\end{eqnarray}
which contains $2D+1$ parameters $\{ W^{f, k}_i (\ell) \}^{2D}_{i=0}$.
For an arbitrary filter, there are $K-1$ such components (one for each dimension $k\in\{1, 2, \dots, K-1 \}$).
Hence, for a single filter, the total number of parameters for all intermediate $k$-dimensional components combined is $(K-1)(2D+1)$.
To that end,for all $k=0, \dots, K$, we see that one filter contains $2(D+1) + (K-1)(2D+1)$ parameters.
Finally, because this holds for any filter in any layer, $F [2(D+1) + (K-1)(2D+1)] L$ total simplicial convolutional parameters.

\end{document}